\documentclass[10pt,twocolumn,letterpaper]{article}

\usepackage{cvpr}              

\usepackage[dvipsnames]{xcolor}

\definecolor{cvprblue}{rgb}{0.21,0.49,0.74}
\usepackage[pagebackref,breaklinks,colorlinks,citecolor=cvprblue]{hyperref}
\usepackage{comment}
\usepackage{changepage}
\usepackage{indentfirst}
\usepackage{algorithm}
\usepackage{algpseudocode}
\usepackage{multirow}
\usepackage{graphicx}
\usepackage{pifont}
\usepackage{makecell}
\usepackage{rotating}
\usepackage[accsupp]{axessibility}
\def\paperID{17} 
\def\confName{CVPR}
\def\confYear{2024}

\title{MambaPupil: Bidirectional Selective Recurrent model \\for Event-based Eye tracking}

\author{Zhong Wang, Zengyu Wan, Han Han, Bohao Liao, Yuliang Wu, \\Wei Zhai\footnotemark, Yang Cao, Zheng-jun Zha\\
University of Science and Technology of China\\
\{\tt\small wangzhong@mail., wanzengy@mail., hanh@mail., liaobh@mail., tronliang@mail., \\
\tt\small wzhai056@, forrest@, zhazj@\}ustc.edu.cn
}

\begin{document}
\maketitle

\renewcommand{\thefootnote}{\fnsymbol{footnote}}
\footnotetext{*Corresponding author.}

\begin{figure*}[h]
    \centering
    \includegraphics[width=.95\linewidth]{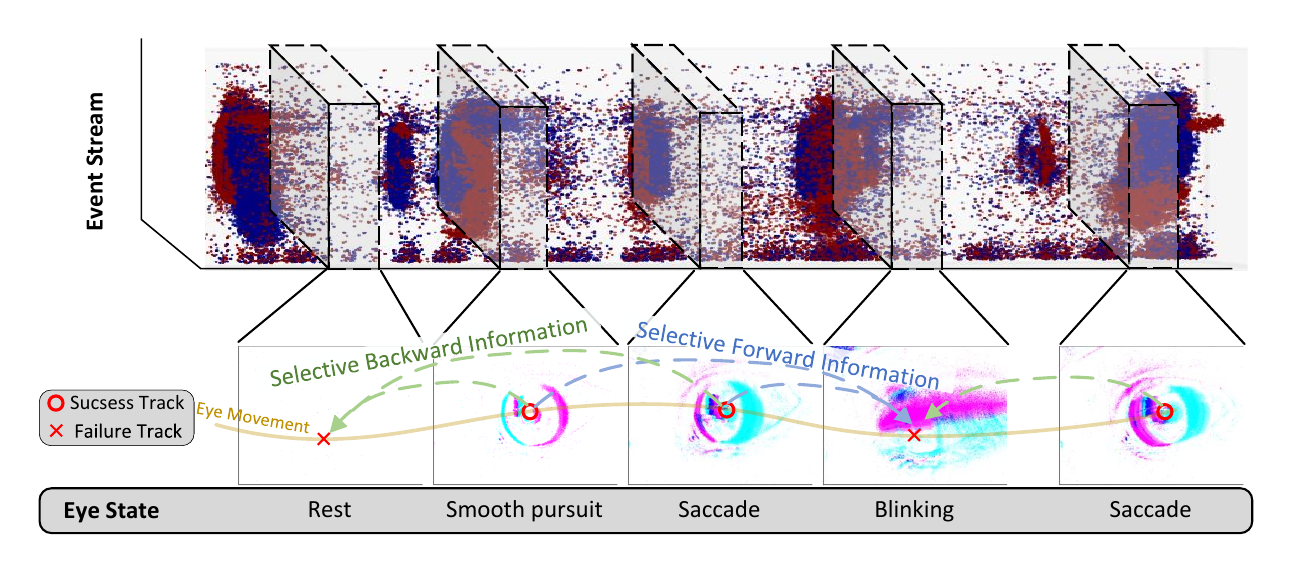}
    \caption{The diversity of eye movements, including blinking, saccades, resting, etc., poses significant challenges for accurate and stable eye tracking. To address these challenges, leveraging contextual temporal information becomes crucial in assisting with eye localization. The historical and future data of eye movements can help to infer the current position and potential trajectories of the eye.}
    \label{fig:enter-label}
\end{figure*}


\begin{abstract}
    Event-based eye tracking has shown great promise with the high temporal resolution and low redundancy provided by the event camera. However, the diversity and abruptness of eye movement patterns, including blinking, fixating, saccades, and smooth pursuit, pose significant challenges for eye localization. To achieve a stable event-based eye-tracking system, this paper proposes a bidirectional long-term sequence modeling and time-varying state selection mechanism to fully utilize contextual temporal information in response to the variability of eye movements. Specifically, the \textbf{MambaPupil} network is proposed, which consists of the multi-layer convolutional encoder to extract features from the event representations, a bidirectional Gated Recurrent Unit (GRU), and a Linear Time-Varying State Space Module (LTV-SSM), to selectively capture contextual correlation from the forward and backward temporal relationship. Furthermore, the Bina-rep is utilized as a compact event representation, and the tailor-made data augmentation, called as \textbf{Event-Cutout}, is proposed to enhance the model's robustness by applying spatial random masking to the event image. The evaluation on the ThreeET-plus benchmark shows the superior performance of the MambaPupil, which secured the 1st place in CVPR'2024 AIS Event-based Eye Tracking challenge.
\end{abstract}    
\section{Introduction}
\label{sec:intro}

Event-based eye tracking aims to precisely predict the pupil's spatial location at any moment based on the event camera signal collected within the human eye area. It has demonstrated great importance in the field of human-computer interaction (HCI)\cite{hci}, and the eye-tracking component promises the potential to be a significant functionality of future virtual reality/augmented reality (VR/AR) devices\cite{vr}. The traditional eye tracking systems rely on high-speed cameras \cite{high-speed-camera}, which are power-intensive and demanding for light-weight design. In contrast, the event camera provides the possibility to achieve a practical eye-tracking system with its low power consumption and high temporal resolution.

As the fastest moving organ in the human body, the state of the pupil constantly changes drastically and abruptly, which poses several hard challenges for eye tracking: 1) Loss of target due to blinking: When a person blinks, the eyelid completely covers the eyeball, leading to a brief loss of the pupil's state. Moreover, blinking generates numerous irrelevant events, which may cause non-negligible fluctuations in the model's predictions in a short period; 2) Sparse events during eye resting: As event cameras generate signals when brightness changes occur, the event information produced when the eye is at rest is extremely sparse. When the event density is insufficient to support accurate predictions, the results typically deviate and exhibit noticeable jitters; 3) Interference from other objects: These objects usually refer to glasses, eyelashes, and reflections on irises. The former generates signals when the head moves, while the latter two produce signals during eye movements. These signals are often detrimental to pupil tracking, especially the presence of glasses, which can cause significant and prolonged misalignment of the predictions. 

To resolve the above challenges, digging out the contextual temporal information is obviously the key factor. Recent works try to extract the temporal correlation of pupil movement by the Recurrent Neuron Network (RNN), such as LSTM\cite{3et}, to realize stable tracking. Additionally, there exist some attempts to employ a Spiking Neuron Network (SNN)\cite{snn-et} to track, which is tailored for event signals' asynchronous and focuses on more fine temporal relationships. However, in these works, the temporal direction is considered unidirectional, and each timestamp is taken into account equally, rendering too simple a temporal model and prone to failure in challenging eye tracking.

To fully extract and utilize the temporal correlation of pupil movement, this paper attempts to model the temporal relationship bidirectionally and selectively and proposes the \textbf{MambaPupil} model for stable eye tracking. Specifically, the model utilizes the stacks of the CNN layer to extract event spatial information, which is then fed into a Dual Recurrent Module to extract the specific pupil position by contextual temporal modeling, where the bidirectional GRU is firstly employed to extract contextual information, which is input into the Linear Time-Varying State Space Module (LTV-SSM) to establish the hidden state of eye movement patterns\cite{mamba}. 

To further improve the model's generalization performance, the Bina-rep, an event spatial-temporal binarization representation, is utilized as network input\cite{bina-rep}. This method transforms multiple event frames accumulated within a time window into binary event graphs of specified digits, significantly reducing the input scale and avoiding interference caused by noisy events. A tailor-made data augmentation technique, \textbf{Event-Cutout}, is proposed to enhance the robustness of spatial localization by spatial random masking\cite{masked-event-modeling}. Experiments on challenging eye-tracking dataset demonstrate that the proposed method achieves accurate and stable tracking results even on hard samples. Moreover, due to the concise design of the model, it has low training costs while maintaining extremely fast inference speeds.

In summary, our main contributions are as follows:

\begin{adjustwidth}{2em}{0pt}
\begin{itemize}
  \item A novel framework, MambaPupil, is proposed to model the temporal relationship bidirectionally and selectively for accurate and stable eye tracking.
  \item The Dual Recurrent Module in MambaPupil is designed to utilize the bidirectional GRU module to rich the temporal contextual information by bidirectional temporal modeling and the LTV-SSM module to selectively cast importance into the valid eye motion stage by input-dependent weight adjusting.
  \item The tailored data augmentation technique, Event-Cutout, is proposed by random spatial masking, endowing the network strong robustness under challenging senses.
  \item Evaluation on challenging eye tracking dataset demonstrates the superior performance of the proposed approach.
\end{itemize}
\end{adjustwidth}

\section{Related Work}
\label{sec:related_work}

\begin{figure*}[htbp]
  \centering
  \includegraphics[width=.95\hsize]{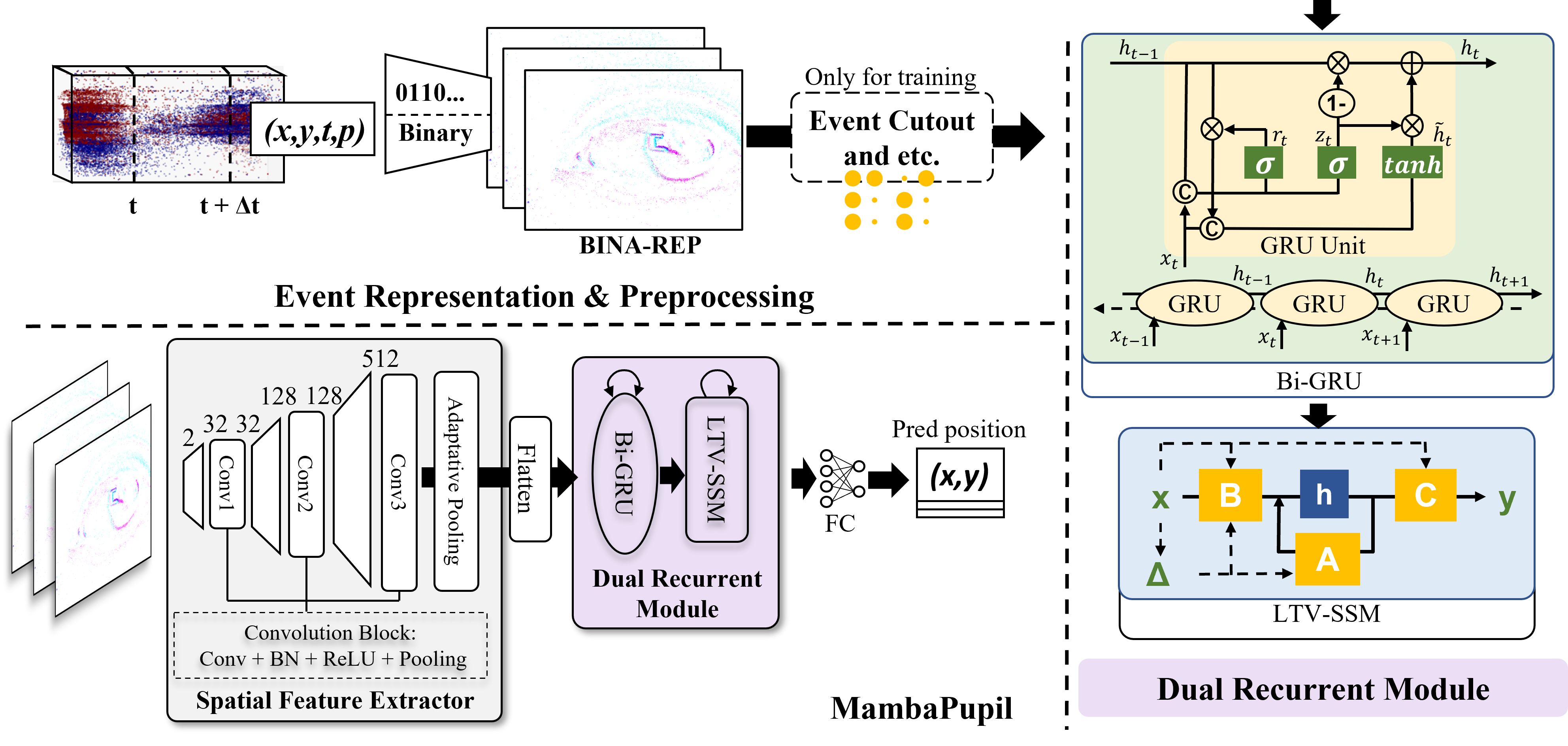}
  \caption{\textbf{Top}: Event streams are transformed into Bina-rep representations and enhanced only during the training phase. \textbf{Bottom}: MambaPupil framework is composed of the Spatial Feature Extractor, Dual Recurrent Module, and the Fully Connected layer as the final classifier. The Spatial Feature Extractor consists of stacked convolutional blocks. \textbf{Right}: The Dual Recurrent Module consists of two temporal modeling units, Bi-GRU and LTV-SSM. The Bi-GRU extracts the complete contextual information through the bi-directional temporal information flow process, and LTV-SSM achieves selective temporal state encoding through content-dependent parameter tuning.}
  \label{fig:method}
\end{figure*}

\subsection{Eye Tracking}

In eye-tracking tasks, model-based methods developed according to the structure of the eye have achieved very high accuracy\cite{et2}. These methods extract prominent geometric features of the eye from images or utilize optimization techniques to perform 3D physical modeling of the eye based on corneal and retinal reflection patterns\cite{glint-et,glint-et2, zhai2024background}. However, these methods require additional equipment and frequent calibration, and they are sensitive to ambient lighting conditions. Moreover, they are susceptible to factors such as pupil color, eyelid occlusion, and eyelash occlusions\cite{phy-et, zhai2022exploring}. Appearance-based End-to-end learning methods can directly infer the position of the pupil from features of input images extracted by CNN\cite{im-et1,im-et2,im-et3, zhai2023exploring, zhai2022one, im-et4}. However, the tracking speed of these methods is limited by the camera's frame rate, making it challenging to achieve sub-millisecond time resolution on wearable devices.

Event-based methods, due to the asynchronous nature of the signals, have the potential to address the time resolution issues of end-to-end models\cite{event-survey}. Early event-based eye tracking approaches used modeling methods as a backbone, fitting image, and event frames to parameters of the 3D physical model\cite{model-based-et}. While this method achieves impressive model accuracy, it is still dependent on image information and significantly impacted by sensor noise. Recent purely event-based eye tracking algorithms can be divided into two directions: those based on Spiking Neural Networks (SNN) and those based on event frames. SNN-based networks\cite{snn-et}, composed of neural cell assemblies, can fully utilize the asynchronous nature of events and are lightweight. The latter converts events into event frames with a fixed rate and extracts features related to pupil position using processing techniques similar to grayscale images\cite{3et}. Various RNN structures, such as LSTM, are commonly used to capture the temporal information of events. Aligning with existing event frame-based methods, we further improve performance through novel network design and data representing techniques.

\subsection{State Space Model (SSM)}

The state space model (SSM) is a concept based on control theory. It introduces a state vector and its update matrices between input and output variables, aiming to model and analyze systems more finely, allowing for more precise control of the target object. Gu et al. introduced it into the field of deep learning in \cite{s4}. They first proposed the S4 model, which captures the intrinsic features of specific tasks through the analysis of Linear Time-Invariant (LTI) systems. Subsequently, they further improved the structural simplicity and performance by introducing the S4D and S5 models. These models are utilized in a way similar to other RNN varieties to capture dependencies over different time ranges. Due to the limitations of time-invariant systems in handling all cases, in \cite{mamba}, they proposed the time-variant S6 model, which not only allows more parameters to be correlated with inputs but also introduces selection strategies and performance optimizations at the hardware level. In our network architecture, after generating hidden states with temporal context information using GRU, we fit a time-variant SSM to model these states, achieving excellent results in eye-tracking tasks. Recently, there have also been some works applying SSM to event-based computer vision\cite{Zubic_2024_CVPR}.
\section{Methodology}
\label{sec:methodology}

In this section, we describe the preprocessing methods of event data and augmentation techniques in Sect.~\ref{section:event-representation}. Then, we explain the structure of the proposed network in Sect.~\ref{section:network}. Finally, we introduce the loss function in Sect.~\ref{section:loss}. The whole framework is visualized in Fig.~\ref{fig:method}.

\subsection{Event Processing}
\label{section:event-representation}

Event cameras asynchronously generate event data at each pixel independently. When the logarithmic change in brightness at a pixel position $(x, y)$ surpasses the threshold, an event is triggered as :
\begin{align}
\label{equ1}
    log\mathcal{I}(x,y,t)-log\mathcal{I}(x,y,t-\delta t)=p\textit{C},
\end{align}
where $\mathcal{I}$ represents the scene brightness, $C$ is the set threshold, and $\delta t$ is the timestamp of the event, measured in microseconds. $p \in \{ -1, 1\}$ represents the polarity of the event change, indicating either decrease or increase in brightness. 

Such event points are motion-triggered in the spatiotemporal domain, forming the event stream and we transform a fixed period of event  into the specific event representation for eye tracking. Next, we will detail our event representation, Bina-rep and the corresponding data augment techniques introduced in training phase.

\begin{figure}[tp]
  \centering
  \includegraphics[width=.95\linewidth]{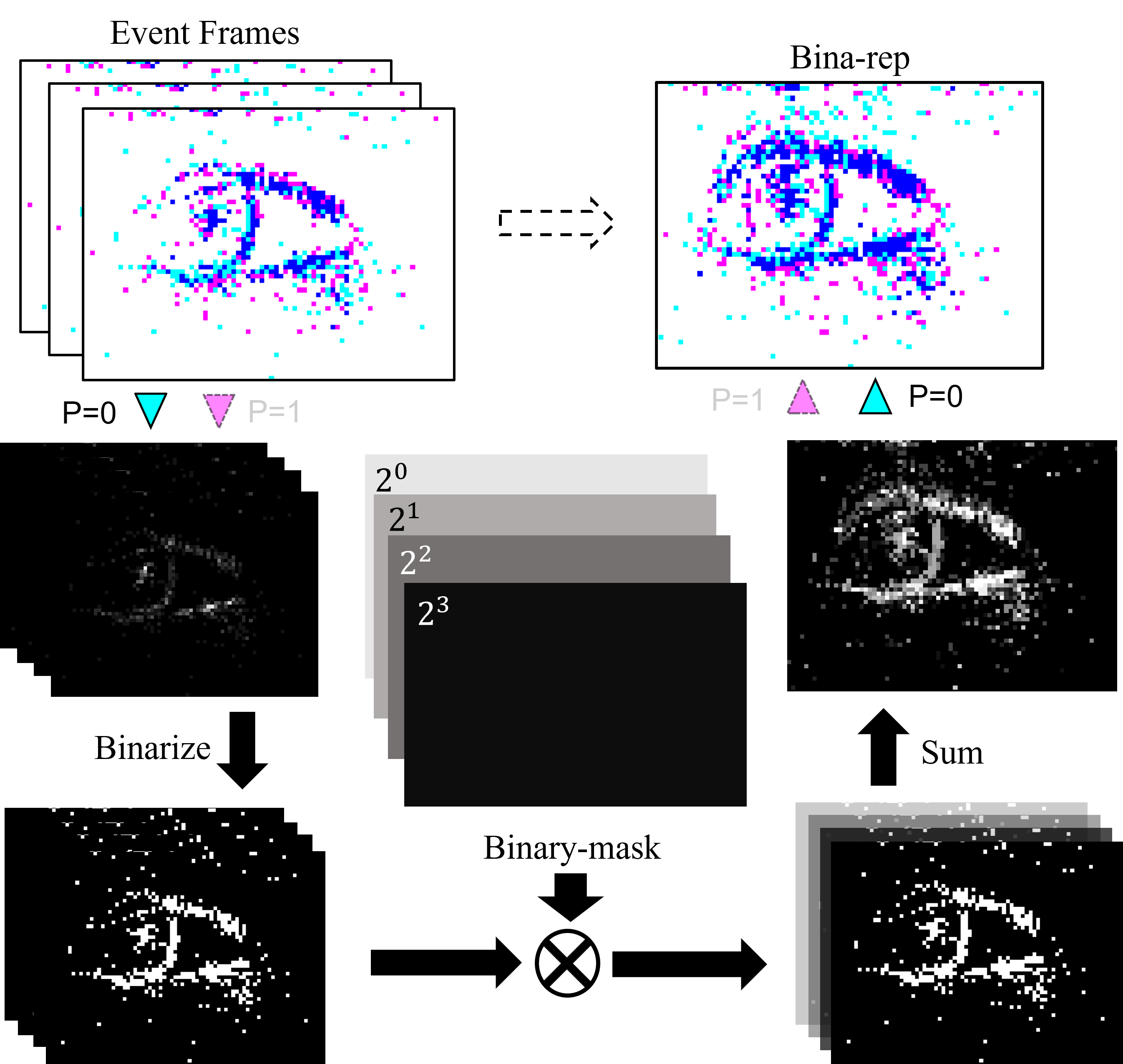}
  \caption{The Bina-rep is generated by aggregating binary-masked event frames. The routines for each polarity of event are the same and will be concatenated at last.}
  \label{fig:bina-rep}
\end{figure}

\textbf{Bina-rep Representation of Event data.} Due to the similarity of spatial shape of the eye, we adopt a binarization quantization method proposed by Barchid et al.\cite{bina-rep}, which converts the event over a period of time $\Delta t$ into a representation format of N-bit map, as shown in Fig.~\ref{fig:bina-rep}. Specifically, it first generates a spatially binarized event clip based on the event frames aggregated and accumulated over a period of time, after which it encodes each frame in the clip using a temporal binarization mask. And then, we accumulate the spatio-temporally binarized encoded clip to form the final Bina-rep $B_e$. This approach brings two benefits: 1) it reduces storage and computational costs, improving the training and inference speed of the network. The space occupied by the Bina-rep representation for the same time-length of event data is one divided by the $n_{bins}$ in the standard event frame representation; 2) it reduces the impact of noise and redundant information on prediction results to some extent. In this representation method, isolated events are less pronounced.

\begin{figure}[tp]
  \centering
  \includegraphics[width=\hsize]{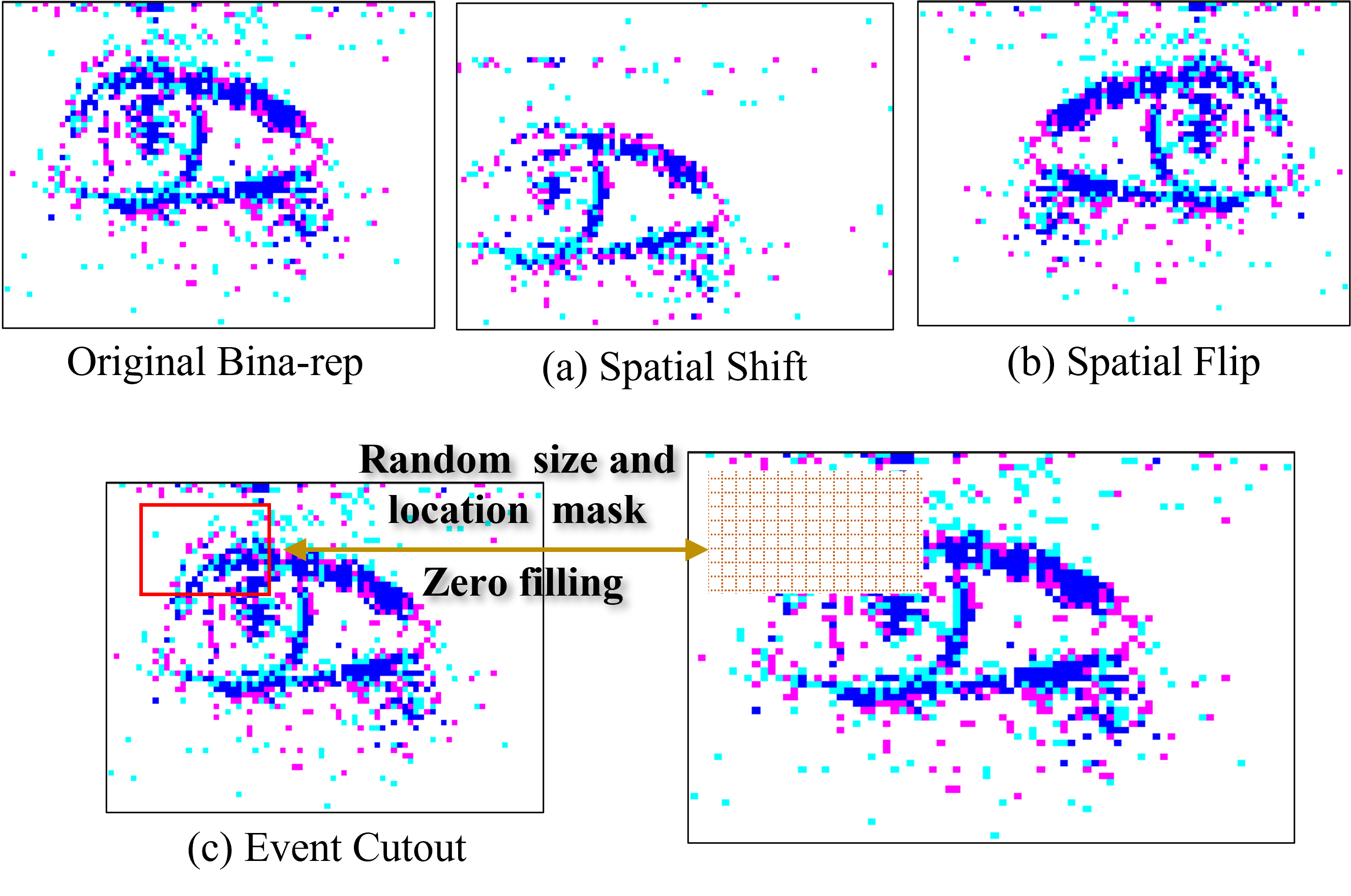}
  \caption{Various data enhancement methods are adapted to improve the robustness, including spatial flip, spatial shift, event-cutout and etc. Event-Cutout is a spatial enhancement technique suitable for event-based eye tracking, which sets the pixels within random rectangular box to zero to simulate potential external interference, \textit{e.g.}, eye blinking.}
  \label{fig:cutout}
\end{figure}

\textbf{Event-Cutout and Other Data Augment Techniques.} To enhance the algorithm's generalization ability when dealing with challenge scenes, \textit{e.g.} eye blinking, we utilize several event data enhancement techniques in training phase, including spatial flip, spatial shift, event-cutout and etc. These data enhancement techniques are visualized Fig.~\ref{fig:cutout}. And the event-cutout is a tailor-made data enhancement technique for eye tracking, of which the core idea is randomly masking spatial regions, forcing the network to learn global features and avoid getting trapped in local minutiae\cite{masked-event-modeling}. The event-cutout is designed by randomly sampling the height and width value as the size of spatial mask and also randomly sampling the x, y coordinate as the location of the mask location in original Bina-rep. The region in this mask will be filled into 0, indicating the event in this region will be free. This technique helps to generate the spatial interference in training phase and is particularly effective in scenarios such as blinking or wearing glasses, which hinders the model to focus on valid eye region.

\subsection{Architecture of Proposed Network}
\label{section:network}

In this section, we will introduce the network architecture demonstrated in Fig.~\ref{fig:method}. Overall, our network consists of two main parts: the Spatial Feature Extractor employs a convolutional encoder to extract spatial semantic features from the Bina-rep, and the Dual Recurrent Module utilizes a Bi-GRU and LTV-SSM to extract temporal constraints bidirectionally and selectively for predicting the pupil's location.

\textbf{Spatial Feature Extractor.} The input Bina-rep is firstly transported into the Spatial Feature Extractor, which consists of a series of convolution blocks to extract spatial pattern $x_t$ at time t. Each convolution block is designed as following,
\begin{align}
    x_t = Pool(ReLU(BN(Conv(B_e)))),
\end{align}
where the $Conv$ is the 2d convolution operator, $BN$ is the batch normalization layer, $ReLU$ is the relu activation function and the $Pool$ is the spatial pooling layer. To obtain spatial correlation of a broader range, larger convolutional kernels (7 or 5) are employed. After passing through three convolutional blocks, the features are fed into the global spatial pooling layer for aggregation. Subsequently, A spatial dropout\cite{dropout} is adopted in the training phase, and the results are fed into the Dual Recurrent Module.

\textbf{Dual Recurrent Module.} In the Dual Recurrent Module, two consecutive recurrent modules are sequentially employed to extract relevant temporal information bidirectionally and selectively. The first module consists of a bidirectional Gated Recurrent Unit (GRU)\cite{gru}. It computes update and reset gate output $r_t$, $z_t$, based on inputs $x_t$ and the previous time step's hidden state $h_{t-1}$, controlling the memory and forgetfulness of the hidden state. Compared with Long Short-term Memory (LSTM) network, the GRU is easier to train and less prone to issues like gradient vanishing and explosion. However, using a unidirectional network for the temporal segment may lead to non-ideal predictions at one end because of information insufficiency. To tackle this, bidirectional GRU units are employed to enhance prediction accuracy at both ends of segments, in which the sequential input will be forward and backward passed through the GRU, and the bi-directional hidden states will be concatenated as the final output. The Bi-GRU is employed as follows:

\begin{align}
    z_t &= \sigma(W_z \cdot [h_{t\pm1}, x_{t}]), \\
    r_t &= \sigma(W_r \cdot [h_{t\pm1}, x_{t}]), \\
    \tilde{h}_t &= tanh(W\cdot[r_t * h_{t-1}, x_t]), \\
    h_{t\{f,b\}} &= (1 - z_t) * h_{t\pm1} + z_t * \tilde{h}_t,
\end{align}

The $W_z, W_r, W$ stand for the update, rest and output gate weight respectively, $\sigma$ is the sigmoid activation function, and $h_{t\{f,b\}}$ is the forward or backward hidden state. We will concatenate the forward and backward state as the bidirectional temporal information $h_t$.

After extracting features in the previous stage through Bi-GRU, we feed the hidden states of each time step into the LTV-SSM. We leverage the State Space Model (SSM)\cite{mamba} to model the behavior patterns of eye movements selectively to cast more attention into the valid phase. Here, we first introduce the state space model, which is formalized as follows:
\begin{align}
\label{equ2}
    \Delta x &= Ax + Bu, \\
    y &= Cx + Du,
\end{align}
where \( x \), \( y \), and \( u \) represent the hidden state, the output and control variables of the system, respectively, A is the state (or system) matrix, B the input matrix, C the output matrix, and D the feedforward matrix. When implementing this method as linear time varying system in deep networks, we follow the \cite{mamba} strategy to treat D as fixed network parameters, and matrices A, B and C are influenced by the input sequence. Specifically, during the discretization process, we directly construct \( \Delta \), \( B \), and \( C \) from the input sequence \( x \), while learning the parameters of the constructor during the training process.
\begin{align}
    \Delta, B, C &= Linear(x), \\
    \Delta A &= exp(\Delta * A), \\
    \Delta B &= \Delta * B,
\end{align}

Before feeding the hidden states into the SSM, root mean square normalization (RMSNorm) is applied to enhance computational efficiency. Additionally, outside the entire SSM module, we apply residual connections to mitigate the adverse effects of network complexity. Therefore, the computational process of the LTV-SSM module is as follows:
\begin{align}
    x_{t}^{'} &= RMSNorm(x_t), \\
    h_t &= \Delta A * h_{t-1} + \Delta B * x_{t}^{'}, \\
    y_t &= C * h_t + D * x_t^{'} + x_t,
\end{align}

\subsection{Loss Function for Training}
\label{section:loss}

When computing the loss function, we use the mean square error between the predicted results and the labels, averaged over each segment to avoid bringing extra complexity. Below is our loss calculation formula,
\begin{align}
    Loss = \sqrt{\frac{1}{L}\sum_{i=1}^{L}(y_{i,pred} - y_{i,label})^2} ,
\end{align}
where $y_{i,pred}$ and $y_{i,label}$ represent the predicted results and training labels, respectively, for the $i$-th sample in the segment, and $L$ denotes the length of each segment involved in training.

\begin{table*}[tp]
\centering

\begin{tabular}{lc|ccccc}
\toprule
Method & train stride & $p_{5}$ & $p_{10}$ & $p_{15}$ & $p_{error}$ & validation loss \\
\midrule
 & 30 & 0.610 & 0.856 & 0.931 & 5.90 & 0.0763 \\
CNN-GRU\cite{kaggle,wang2024ais_event} & 10 & 0.605 & 0.832 & 0.928 & 5.90 & 0.0761\\
 & 5 & 0.615 & 0.844 & 0.931 & 5.90 & 0.0737\\
\hline
 & 30 & 0.788 & 0.936 & 0.971 & 3.92 & 0.0542\\
CB-ConvLSTM\cite{3et} & 10 & 0.858 & 0.968 & \underline{0.989} & 5.80 & 0.0427\\
 & 5 & 0.792 & 0.848 & 0.981 & 3.82 & 0.0543\\
\hline
 & 30 & 0.903 & 0.971 & 0.985 & 2.77 & 0.0396 \\
MambaPupil(Ours) & 10 & \underline{0.935} & \underline{0.976} & 0.987 & \underline{2.35} & \underline{0.0322} \\
& 5 & \textbf{0.937} & \textbf{0.984} & \textbf{0.990} & \textbf{2.03} & \textbf{0.0287} \\
\bottomrule
\end{tabular}

\caption{
Comparison with other methods: $p_n$ denotes prediction accuracy within n pixels, where higher values indicate better performance, and $p_{error}$ represents the average prediction Euclidean distance, where lower values are preferred. We compare models trained with different training strides. It can be observed that our method consistently outperforms other methods across various training strides.
}
\label{tab:methods}
\end{table*}

\section{Experiments}
\label{sec:experiments}

\subsection{Datasets and Preprocessing}

We perform experiments on the EET+ dataset for training and evaluation\cite{kaggle,wang2024ais_event}. This dataset comprises 13 recorded scenarios, each containing 2-6 segments, all with an event resolution of 640x480. The motions of the human eye in these segments can be roughly categorized into five classes: random movement, saccades, reading text, blinking, and smooth pursuit. The labels for pupil positions were performed at 100Hz, and we utilized 3/4 of the data as the training set, 1/4 as the validation set, and predicted labels at a sampling rate of 20Hz following the \cite{kaggle} setting. The evaluation was assessed by the Euclidean distance between the predicted pupil center and the ground truth and the successful rate in which distances shorter than p pixels are regarded as successful. In this context, $p_{5}$, $p_{10}$, and $p_{15}$ represent p = 5, 10, and 15,  are adopt to evaluate the performance.

\subsection{Implementation Details}

The proposed method was implemented using PyTorch\cite{pytorch}, and the experiments were conducted on a single NVIDIA RTX2080Ti or GTX1080Ti GPU. The training epoch was set to 1000 with a batch size of 32. All networks utilized the Adam optimizer and the Cosine Annealing Warm Restart scheduler\cite{cosanl}, starting with a learning rate of 0.002. To improve training efficiency, the resolution of the event and label was downsampled to 80x60. The Spatial Feature Extractor in MambaPupil is composed of three convolution blocks with output channels of 32, 128, and 512, respectively, and the kernel sizes in blocks were 7, 5, and 5. As for the Bina-rep, the bit number was set to 4 empirically. Each training sequence had a length of 45 and was trained with a step size of 5; different training strides had a significant impact on the results, with smaller step sizes yielding superior performance due to generating more hard sequences, albeit at the cost of increasing training time by multiples. The following ablation and comparison experiments will mainly conducted at the train stride of 10 for efficient consideration. The data augmentation methods were conducted on 50\% of the training data. As for labels for training, The dataset provides pupil position labels at 640x480@100Hz. To align with our data processing pipeline, we downsampled the time resolution to 20Hz, and spatial position of both predictions and labels are normalized. 

\begin{table}[t]
\centering

\begin{tabular}{lccc}
\toprule
Model & Params & Flops & Train time \\
\midrule
CNN-GRU & 37.23M & \underline{1.9T} & \underline{1h42m} \\
CB-ConvLSTM & \textbf{417.17K} & \textbf{1.09T} & 2h12m\\
MambaPupil & \underline{8.59M} & 2.61T & \textbf{1h31m} \\
\bottomrule
\end{tabular}

\caption{
Evaluated on a single RTX2080Ti for 1000 epoches. Different train time for 1000 epoches in various train strides. Batch size and train stride is fixed at 32 and 10. 
}
\vspace{-0.05in}
\label{tab:ts}
\end{table}

\subsection{Comparision with Other Methods}

In this section, we will compare the training performance of different methods on this dataset. This includes a vanilla baseline based on CNN and GRU\cite{kaggle}, as well as the state-of-the-art method based on multi-layer LSTM proposed in \cite{3et}. From Tab. \ref{tab:methods}, we can observe that our method exhibits superior performance than existing methods, achieves the 9.0\%/1.5\%/0.1\% improvement in  $p_5$, $p_{10}$, and $p_{15}$, and reduces the error distance to 2 pixels. Our network achieves excellent performance while effectively controlling the scale of parameters and computational operations. As shown in the Tab. \ref{tab:ts}, we need shorter training time compared to other methods. 

\subsection{Ablation Study}

In this section, we will investigate the impact of different factors on the results in detail, including settings of the GRU and the LTV-SSM in the Dual Recurrent Module,  Event representations, and data augmentation techniques.

\textbf{Bi-GRU \& LTV-SSM in Dual Recurrent Module}: To verify the effectiveness of the Dual Recurrent Module design, we attempt to remove the Bi-GRU or LTV-SSM in it and observe how the performance changes. The results are shown in Tab. \ref{tab:ablation-stru}, which demonstrates the validness of the Dual Recurrent Module. Specifically, adopting Uni-GRU to replace the Bi-GRU causes the significant tracking error increase in $p_{error}$ and remarkable missing rate increase in $p_{5}$ and $p_{10}$. It is owing to the lack of context information at the start and end positions of each sequence and bidirectional information helps to capture the valid pupil state to cope the difficult scenarios. In further, removing the LTV-SSM results in an increase in the prediction error by nearly 0.1 pixel, and affects the stability of the predictions by doubling the missing rate within 5 and 10 pixels. It dues to the LTV-SSM exploit the contextual relationship selectively, which helps to cast more attention in informative timestamp, like the smooth pursuit phase and less attention in resting or blinking phase. Thus, it can contribute to the stability of prediction and especially brings the improvement on $p_5$ metric.

\begin{table}[tp]
\centering

\centering
\begin{tabular}{ccc|ccc}
\hline
     \multicolumn{2}{c}{GRU} & \multirow{2}{*}{LTV-SSM}  & \multicolumn{3}{c}{Metrics} \\
    \cline{1-2}
     Uni & Bi                &         &  $p_{5}$ & $p_{10}$ & $p_{error}$                  \\
    \hline
     \ding{51} & \ding{55} & \ding{55} & 0.916 & \textbf{0.981} & \textbf{2.27} \\
     \ding{51} & \ding{55} & \ding{51} & 0.901 & 0.972 & 2.68 \\
     \ding{55} & \ding{51} & \ding{55} & \underline{0.919} & \textbf{0.981}  & \underline{2.29} \\
     \ding{55} & \ding{51} & \ding{51} & \textbf{0.935} & \underline{0.976} & 2.35 \\ 
\hline
\end{tabular}

\caption{
Comparison of different structure designs in the dual recurrent module. We adapted bi-directional and uni- version of GRU, and attempted to remove LTV-SSM module in each instance.}

\label{tab:ablation-stru}
\end{table}

\textbf{Event Representations}: This part evaluates the impact of different event representations. From Tab. \ref{tab:ablation-pre}, it can be seen that the Bina-rep helps to improve the tracking accuracy effectively, especially within a 5-pixel range. It indicates that the Bina-rep contributes to a more stable tracking curve by reducing the influence of other interference factors, e.g., time jitters from the camera. Thus, it can improve the stability of the model, even in the fast eye movement phase.

\begin{table}[tp]
\centering

\begin{tabular}{lccc}
\toprule
Event Representation & $p_{5}$ & $p_{10}$ & $p_{error}$\\
\midrule
Voxel-grid & 0.891 & \underline{0.978} & 2.56 \\
Frame & \underline{0.912} & \textbf{0.990} & \underline{2.42} \\
Bina-rep & \textbf{0.935} & 0.976 & \textbf{2.35} \\
\bottomrule
\end{tabular}
\caption{
Comparison on the different event representations.
}
\label{tab:ablation-pre}
\end{table}

\textbf{Event-Cutout and other data enhancing techniques}: The data enhancement techniques contribute to improving the accuracy of predictions under challenging scenarios, e.g., eye blinking. Adopted augmentation techniques include: 1) Position offsets: It is usually hard to estimate when pupils' position is on the edge of the frame. Shifting spatial positions of samples helps to obtain more edge samples for training, thus enhancing performance in such conditions; 2) Spatial flips: randomly flipping the frame, especially on the Y-axis, helps to understand the motion pattern of eyes in different viewing angles. As it is rare to have eye frames upside down to train the network, such a flip can enhance the generalization of the model; 3) Time shift: We slice the raw event using a fixed time window, and it carries the risk of information loss, which can be alleviated by utilizing a time shift. We conducted ablation studies on both the event cutout method and the above data augmentation techniques; the results are shown in Tab. \ref{tab:ablation-aug}. It can be seen that these data enhancing techniques prove more difficult samples to improve the model's generalization ability, especially the spatial transformation, \textit{e.g}., spatial shift and event-cutout, create the most valid adversarial samples.

\begin{table}[tp]
\centering
\begin{tabular}{lcccc}
\toprule
Augmentations & $p_{5}$ & $p_{10}$ & $p_{error}$\\
\midrule
w/o spatial shift & 0.872 & 0.956 & 3.37 \\
w/o temporal shift & \underline{0.906} & \underline{0.974} & \underline{2.51}\\
w/o spatial flip & 0.888 & 0.961 & 2.90\\
w/o Event-Cutout & 0.895 & 0.959 & 2.83 \\
Full Augmentations & \textbf{0.935} & \textbf{0.976} & \textbf{2.35} \\
\bottomrule
\end{tabular}
\caption{
Comparison on the different data augmentation techniques. }
\label{tab:ablation-aug}
\end{table}

\begin{figure}[t]
    \centering
    \includegraphics[width=\hsize]{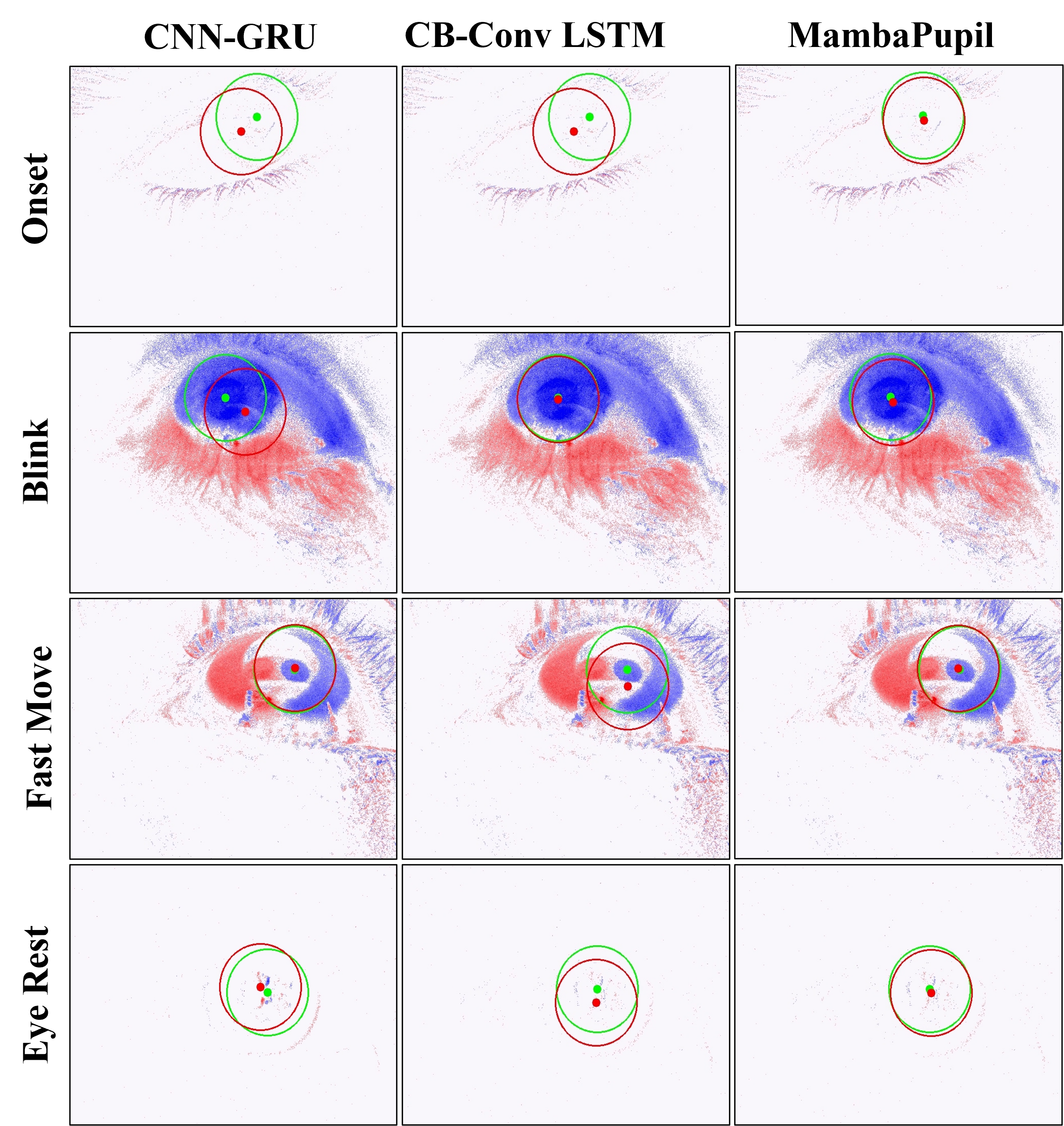}
    \caption{Visual comparison of state-of-the-art model and our MambaPupil in four challenging scenarios, including onset, blink, fast move and eye rest.}
    \label{fig:vis-method}
\end{figure}

\subsection{Qualitative Comparison}
\begin{figure}[t]
    \centering
    \includegraphics[width=\hsize]{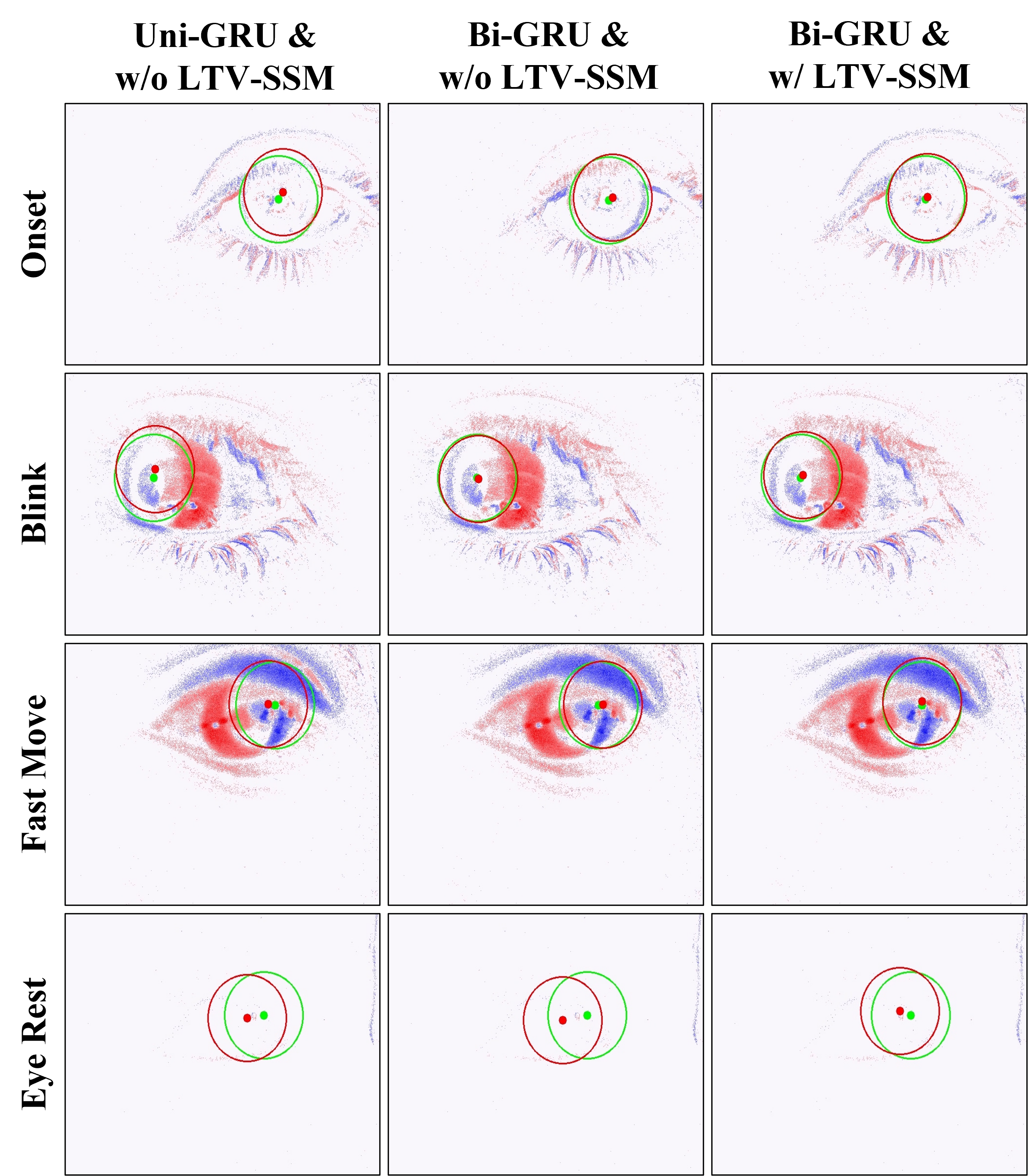}
    \caption{Qualitative comparison of structure design in dual recurrent module.}
    \label{fig:vis-stru}
\end{figure}

In this section, we conduct qualitative comparisons of different models and structure designs in dual recurrent modules to comprehensively verify the effectiveness of the MambaPupil. The visualization results are shown in Fig. \ref{fig:vis-method} and Fig. \ref{fig:vis-stru}. We selected four typical and also challenging scenarios: \textbf{Onset of event slice}: At the initial end of a segment, there is typically a lack of contextual information, which poses a significant disadvantage for RNN networks relying on temporal correlations; \textbf{Blinking and rapid eye movement process}: Blinking generates a large number of interfering event, which can cause the network's prediction results to shift and jitter; \textbf{Eye rest}: Event cameras struggle to capture static objects, resulting in minimal information available to infer pupil position when the eyes are still. These scenarios can cause poor performance in predictions of existing methods. However, the visual results proved that the MambaPupil overcomes these challenges and still achieves high as well as stable tracking accuracy compared to the CNN-GRU and CB-ConvLSTM model. To delve deeper into the roles played by different modules in the network, we sequentially discarded the LTV-SSM module and bidirectional design in Bi-GRU and observed a remarkable performance loss in prediction in Fig. \ref{fig:vis-stru}.

\section{Conclusion}
\label{sec:conclusion}

In this paper, we analyzed the eye tracking task based on pure event data and proposed MambaPupil framework that integrates Bi-GRU and LTV-SSM modules to fully obtain temporal correlations, enabling accurate and fast-tracking of pupil position. Furthermore, with the Bina-reps and delicately designed data enhancement, we outperform the state-of-the-art model and achieve excellent results even in challenging scenarios. We believe this approach can provide valuable insights for other motion-tracking tasks. For future work, we aim to fully leverage the asynchronous nature of event data and further explore the generalization potential of state space models in other event-based tasks.
\\ \\
\noindent\textbf{Acknowledgments.} This work is supported by the National Natural Science Foundation of China (NSFC) under Grants 62225207 and 62306295.
{
    \small
    \bibliographystyle{ieeenat_fullname}
    \bibliography{main}
}


\end{document}